%% file: main.tex
\documentclass{article}

\usepackage{microtype}
\usepackage{float}
\usepackage{graphicx}
\usepackage{booktabs} 
\usepackage{amsmath}
\usepackage{amssymb}
\usepackage[export]{adjustbox}
\usepackage{caption}
\usepackage{subcaption}
\usepackage{xcolor}
\usepackage{placeins}
\usepackage{multicol}

\usepackage{tikz}
\usetikzlibrary{arrows, arrows.meta, shapes, shapes.misc, calc, positioning, decorations.pathreplacing, fit, backgrounds}

\tikzset{bicolor/.style args={#1 and #2}{
  path picture = {
    \tikzset{rounded corners = 0}
    \fill [#1] (path picture bounding box.west)
      rectangle (path picture bounding box.north east);
    \fill [#2] (path picture bounding box.west)
      rectangle (path picture bounding box.south east);
    \draw [densely dotted] (path picture bounding box.west) -- (path picture bounding box.east);
}}}

\usepackage{hyperref}
\usepackage[capitalize]{cleveref}

\usepackage[accepted]{icml2019}

\newcommand{\x}{\mathbf{x}}
\newcommand{\y}{\mathbf{y}}
\newcommand{\z}{\mathbf{z}}

\definecolor{vll-orange}{HTML}{E37238}
\definecolor{vll-green}{HTML}{96BF0D}
\definecolor{vll-dark}{HTML}{464646}
\definecolor{vll-light}{HTML}{757575}

\icmltitlerunning{Benchmarking Invertible Architectures on Inverse Problems}

\begin{document}

\twocolumn[
\icmltitle{Benchmarking Invertible Architectures on Inverse Problems}



\icmlsetsymbol{equal}{*}

\begin{icmlauthorlist}
\icmlauthor{Jakob Kruse}{vll}
\icmlauthor{Lynton Ardizzone}{vll}
\icmlauthor{Carsten Rother}{vll}
\icmlauthor{Ullrich K\"othe}{vll}
\end{icmlauthorlist}

\icmlaffiliation{vll}{Visual Learning Lab, Heidelberg University, Germany}

\icmlcorrespondingauthor{Jakob Kruse}{jakob.kruse@iwr.uni-heidelberg.de}

\icmlkeywords{Machine Learning, ICML}

\vskip 0.3in
]



\printAffiliationsAndNotice{}  

\begin{abstract}
Recent work demonstrated that flow-based invertible neural networks are promising tools for solving ambiguous inverse problems.
Following up on this, we investigate how ten invertible architectures and related models fare on two intuitive, low-dimensional benchmark problems,
obtaining the best results with coupling layers and simple autoencoders.
We hope that our initial efforts inspire other researchers to evaluate their invertible architectures in the same setting and put forth additional benchmarks, 
so our evaluation may eventually grow into an official community challenge.
\end{abstract}

\vspace{-1.5em}\section{Introduction}
\label{intro}
\vspace{-0.25em}%

Both in science and in everyday life, we often encounter phenomena that depend on hidden properties $\x$, which we would like to determine from observable quantities $\y$.
A common problem is that many different configurations of these properties would result in the {\em same} observable state,
especially when there are far more hidden than observable variables.
We will call the mapping $f$ from hidden variables $\x$ to observable variables $\y = f(\x)$ the \textit{forward process}.
It can usually be modelled accurately by domain experts.
The opposite direction, the \textit{inverse process} $\y \rightarrow \x$, is much more difficult to deal with.
Since $f^{-1}(\y)$ does not have a single unambiguous answer, a proper inverse model should instead estimate the full \textit{posterior} probability distribution $p(\x \!\mid\! \y)$ of hidden variables $\x$ given the observation $\y$.

Recent work \cite{ardizzone2018analyzing} has shown that flow-based invertible neural networks such as RealNVP \cite{dinh2016density} can be trained with data from the forward process,
and then used in inverse mode to sample from $p(\x \!\mid\! \y)$ for any $\y$.
This is made possible by introducing additional latent variables $\z$ that encode any information about $\x$ {\em not} contained in $\y$.
Assuming a perfectly representative training set and a fully converged model, they prove that the generated distribution is equal to the true posterior.

\begin{figure}[t!]
\centering
\vspace{-0.5em}%
\includegraphics[width=\columnwidth]{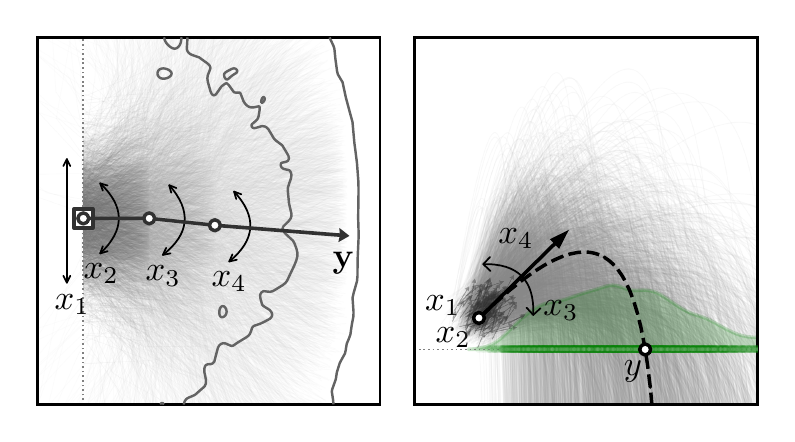}%
\vspace{-1.5em}%
\caption{
Prior distributions of the parameters $\x$ in either benchmark.
\textit{Left:} An articulated arm with three segments is mounted on a rail.
$x_1$ determines the vertical position on the rail and $x_{2 \dots 4}$ determine the angles at the three joints.
$97\%$ of end points of the resulting arms fall within the contour labelled $\y$.
\textit{Right:} An object is thrown upwards and to the right from a starting position $(x_1, x_2)$, at an angle $x_3$ and initial velocity $x_4$.
We observe the locations of impact $y$ where each trajectory hits the ground, i.e.~the $x$ axis.
A green curve shows the density of these positions.
}
\label{fig:inverse-kinematics-prior}\vspace{-1em}
\end{figure}

Interestingly, this proof carries over to all models offering an exact inverse upon convergence.
This poses a natural question: How well can various network types approximate this ideal behavior in practice?
Fundamentally, we can distinguish between {\em hard invertibility}, where the architecture ensures that forward and backward processing are exact inverses of each other (e.g.~RealNVP), and {\em soft invertibility}, where encoder and decoder only become inverses upon convergence (e.g. autoencoders).
The former pays for guaranteed invertibility with architectural restrictions that may harm expressive power and training dynamics, whereas the latter is more flexible but only approximately invertible.

We propose two simple inverse problems, one geometric and one physical, for systematic investigation of the resulting trade-offs.
Common toy problems for invertible networks are constrained to two dimensions for visualization purposes \cite{behrmann2018invertible, grathwohl2018ffjord}.
The 4D problems shown here are more challenging,
facilitating more meaningful variance in the results of different models.
However, they still have an intuitive 2D representation (\cref{fig:inverse-kinematics-prior}) and are small enough to allow computation of ground truth posteriors via rejection sampling,
which is crucial for proper evaluation.
We test ten popular network variants on our two problems to address the following questions:
(i) Is soft invertibility sufficient for solving inverse problems? 
(ii) Do architectural restrictions needed for hard invertibility harm performance? 
(iii) Which architectures and losses give the most accurate results?

\section{Methods}
\label{methods}

\textbf{Invertible Neural Networks (INNs).}\;
Our starting point is the model from \cite{ardizzone2018analyzing}, which is based on RealNVP, i.e.~affine coupling layers.
They propose to use a standard L2 loss for fitting the network's $\y$-predictions to the training data,
\vspace{-0.5em}\begin{align}
    \mathrm{L2}(\y) &= (\y - \y_\mathrm{gt})^2,
    \label{eq:l2}
\end{align}
and an MMD loss \cite{gretton2012kernel} for fitting the latent distribution $p(\z)$ to $\mathcal{N}(\mathbf{0}, \mathbf{I})$, given samples:
\begin{align}
    \mathrm{MMD}(\z) =\,
    &\mathbf{E}_{i,j}[\kappa(\z^{(i)}, \z^{(j)})] - 2 \!\cdot\!
    \mathbf{E}_{i,j}[\kappa(\z^{(i)}, \z_\mathrm{gt}^{(j)})] \,+ \nonumber\\
    &\mathbf{E}_{i,j}[\kappa(\z_\mathrm{gt}^{(i)}, \z_\mathrm{gt}^{(j)})]
    \label{eq:mmd}
\end{align}
With weighting factors $\alpha, \beta$, their training loss becomes
\begin{align}
    \mathcal{L}(\y, \z) &= \mathrm{L2}(\y) + \alpha \cdot \mathrm{MMD}(\z).
    \label{eq:l2-mmd}
\end{align}
\begin{figure}[h!]
    \centering
    \vspace{-1em}%
    \resizebox{0.9\columnwidth}{!}{ \input{figures/INN_model.tex} }%
    \vspace{-0.5em}%
    \label{fig:INN-model}
\end{figure}

We find that it is also possible to train the network with just a maximum likelihood loss \cite{dinh2016density} by assuming $\y$ to be normally distributed around the ground truth values $\y_\mathrm{gt}$ with very low variance $\sigma^2$,
\begin{align}
    \mathcal{L}(\y, \z) =\,
    &\tfrac{1}{2} \cdot \left( \tfrac{1}{\sigma^2} \cdot (\y - \y_\mathrm{gt})^2 +
    \z^2  \right) - \nonumber\\
    &\log \left| \det J_{\x \;\mapsto\, [\y,\,\z]} \right|,
    \label{eq:ml-yz}
\end{align}
and we compare both approaches in our experiments.

\textbf{Conditional INNs}.\;
Instead of training INNs to predict $\y$ from $\x$ while transforming the lost information into a latent distribution,
we can train them to transform $\x$ directly to a latent representation $\z$ \textit{given} the observation $\y$.
This is done by providing $\y$ as an additional input to each affine coupling layer, both during the forward and the inverse network passes. cINNs work with larger latent spaces than INNs and are also suited for maximum likelihood training:
\begin{align}
    \mathcal{L}(\z) =\,
    &\tfrac{1}{2} \cdot \z^2 - \log \left| \det J_{\x \;\mapsto\, \z} \right|,
    \label{eq:ml-z}
\end{align}
\begin{figure}[h!]
    \centering
    \vspace{-1em}%
    \resizebox{\columnwidth}{!}{ \input{figures/cINN_model.tex} }%
    \vspace{-1em}%
    \label{fig:cINN-model}
\end{figure}

\textbf{Autoregressive flows}.\;
Masked autoregressive flows (MAF) decompose multi-variate distributions into products of 1-dimensional Gaussian conditionals using the chain rule of probability \cite{papamakarios2017masked}.
Inverse autoregressive flows (IAF) similarly decompose the latent distribution \cite{kingma2016improved}.
To obtain asymptotically invertible architectures, we add standard feed-forward networks for the opposite direction in the manner of Parallel WaveNets \cite{oord2018parallel} and train with \cref{eq:ml-yz} and a cycle loss:
\begin{align}
    \mathcal{L}(\y, \z, \hat{\x}) =\,
    &\tfrac{1}{2} \cdot \left( \tfrac{1}{\sigma^2} \cdot (\y - \y_\mathrm{gt})^2 +
    \z^2  \right) - \nonumber\\
    &\log \left| \det J_{\x \;\mapsto\, [\y,\,\z]} \right| + \alpha \cdot (\x - \hat{\x})^2
    \label{eq:loss-autoregressive}
\end{align}
\begin{figure}[h!]
    \centering
    \vspace{-2em}%
    \hspace{1em}\resizebox{!}{2cm}{ \input{figures/autoregressive_flow_model.tex} }%
    \vspace{-1em}%
    \label{fig:autoregressive-flow-model}
\end{figure}

\textbf{Invertible Residual Networks}.\;
A more flexible approach is the i-ResNet \cite{behrmann2018invertible}, which replaces the heavy architectural constraints imposed by coupling layers and autoregressive models with a mild Lipschitz-constraint on its residual branches.
With this constraint, the model's inverse and its Jacobian determinant can be estimated iteratively with a runtime vs.~accuracy trade-off.
Finding that the estimated Jacobian determinants' gradients are too noisy
\footnote{While accurate determinants may be found numerically for toy problems, this would not scale and thus is of limited interest.},
we train with the loss from \cref{eq:l2-mmd} instead.
\begin{figure}[h!]
    \centering
    \vspace{-0.5em}%
    \hspace*{3em}\resizebox{!}{2.3cm}{ \input{figures/iResNet_model.tex} }%
    \vspace{-1em}%
    \label{fig:iResNet-model}
\end{figure}

\textbf{Invertible Autoencoders}.\;
This model proposed by \cite{teng2018invertible} uses invertible nonlinearities and orthogonal weight matrices to achieve efficient invertibility.
The weight matrices start with random initialization, but converge to orthogonal matrices during training via a cycle loss:
\begin{align}
    \mathcal{L}(\y, \z, \hat{\x}) &= \mathrm{L2}(\y) + \alpha \cdot \mathrm{MMD}(\z) + \beta \cdot (\x - \hat{\x})^2
    \label{eq:l2-mmd-cycle}
\end{align}
\begin{figure}[h!]
    \centering
    \vspace{-1.5em}%
    \resizebox{!}{2.3cm}{ \input{figures/invAuto_model.tex} }%
    \vspace{-1em}%
    \label{fig:invAuto-model}
\end{figure}

\textbf{Standard Autoencoders}.\;
In the limit of zero reconstruction loss, the decoder of a standard autoencoder becomes the exact inverse of its encoder.
While this approach uses two networks instead of one, it is not subject to any architectural constraints.
In contrast to standard practice, our autoencoders do not have a bottleneck but use encodings with the same dimension as the input (exactly like INNs).
The loss function is the same as \cref{eq:l2-mmd-cycle}.
\begin{figure}[h!]
\centering
\vspace{-1em}%
\hspace{1em}\resizebox{!}{2cm}{ \input{figures/AE_model.tex} }%
\vspace{-1em}%
\label{fig:AE-model}
\end{figure}

\textbf{Conditional Variational Autoencoders}.\;
Variational autoencoders \cite{kingma2013auto} take a Bayesian approach and thus should be well suited for predicting distributions.
Since we are interested in conditional distributions and it simplifies training in this case, we focus on the conditional VAE proposed by \cite{sohn2015cvae}, with loss
\begin{align}
    \mathcal{L}(\boldsymbol\mu_z, \boldsymbol\sigma_z, \hat{\x}) &=
    (\x \!-\! \hat{\x})^2 -\tfrac{1}{2} \alpha \!\cdot\! (1 + \log \boldsymbol\sigma_z - \boldsymbol\mu_z^2 - \boldsymbol\sigma_z).
    \label{eq:elbo-cycle}
\end{align}
\begin{figure}[h!]
    \centering
    \vspace{-2em}%
    \resizebox{!}{2cm}{ \input{figures/cVAE_model.tex} }%
    \vspace{-1em}%
    \label{fig:cVAE-model}
\end{figure}

\textbf{Mixture Density Networks (MDNs)}.\;
MDNs \cite{bishop1994mixture,kruse2020mdn} are not invertible at all, but model the inverse problem directly.
To this end, the network takes $\y$ as an input and predicts the parameters $\boldsymbol\mu_x, \boldsymbol\Sigma_x^{-1}$ of a Gaussian mixture model that characterizes $p(\x \!\mid\! \y)$.
It is trained by maximizing the likelihood of the training data under the predicted mixture models, leading to a loss of the form
\begin{align}
    \mathcal{L}(\boldsymbol\mu_x, \boldsymbol\Sigma_x^{-1}) &=
    \tfrac{1}{2} \!\cdot\! (\x\boldsymbol{\mu}_x^\top \!\cdot\! \boldsymbol{\Sigma}_x^{-1} \!\cdot\! \x\boldsymbol{\mu}_x)
    - \log \lvert \boldsymbol{\Sigma}_x^{-1} \rvert^{\tfrac{1}{2}}.
    \label{eq:ml-mdn}
\end{align}
We include it in this work as a non-invertible baseline.
\begin{figure}[h!]
    \centering
    \hspace{-2em}\resizebox{!}{2cm}{ \input{figures/MDN_model.tex} }%
    \vspace{-1em}%
    \label{fig:MDN-model}
\end{figure}

\section{Benchmark Problems}
\label{benchmark-problems}
We propose two low-dimensional inverse problems as test cases, as they allow quick training, intuitive visualizations and ground truth estimates via rejection sampling.

\begin{table*}[t]
\caption{
Quantitative results for inverse kinematics benchmark, see \cref{sec:experiments}.
The first three columns are averaged over $1000$ different observations $\y^*$.
\textsc{dim$(\z)$} denotes the dimensionality of the latent space.
\textsc{ML Loss} marks models that were trained with a maximum likelihood loss,
while \textsc{$\y$-Supervision} marks models that were trained with an explicit supervised loss on the forward process $\x \rightarrow \y$.
}
\label{tab:kinematics-results}
\begin{center}\begin{small}\begin{sc}
\begin{tabular}{lcccccc}
\toprule
Method & $Err_\mathrm{post}$ (\ref{eq:posterior-mismatch}) & $Err_\mathrm{resim}$ (\ref{eq:re-simulation-error}) & Inference in ms & dim($\z$) & ML Loss & $\y$-Supervision \\
\midrule
INN & 0.025 & 0.015 & 10 & ${\bullet}{\bullet}$ & $\checkmark$ & $\checkmark$ \\
INN (L2 + MMD) & 0.017 & 0.086 & 9 & ${\bullet}{\bullet}$ & & $\checkmark$ \\
cINN & \textbf{0.015} & \textbf{0.008} & 11 & ${\bullet}{\bullet}{\bullet}{\bullet}$ & $\checkmark$ & \\
IAF + Decoder & 0.419 & 0.222 & \textbf{0} & ${\bullet}{\bullet}{\bullet}{\bullet}$ & $\checkmark$ & $\checkmark$ \\
MAF + Decoder & 0.074 & 0.034 & \textbf{0} & ${\bullet}{\bullet}{\bullet}{\bullet}$ & $\checkmark$ & $\checkmark$ \\
iResNet & 0.713 & 0.311 & 763 & ${\bullet}{\bullet}$ & & $\checkmark$ \\
InvAuto & 0.062 & 0.022 & 1 & ${\bullet}{\bullet}$ & & $\checkmark$ \\
Autoencoder & 0.037 & 0.016 & \textbf{0} & ${\bullet}{\bullet}$ & & $\checkmark$ \\
cVAE & 0.042 & 0.019 & \textbf{0} & ${\bullet}{\bullet}$ & & \\
\midrule
\textcolor{gray}{MDN} & \textcolor{gray}{0.007} & \textcolor{gray}{0.012} & \textcolor{gray}{601} & \textcolor{gray}{${\bullet}{\bullet}{\bullet}{\bullet}$} & \textcolor{gray}{$\checkmark$} & \\
\bottomrule
\end{tabular}
\end{sc}\end{small}\end{center}\vspace{-1.5em}
\end{table*}

\begin{figure*}[t]
\centering
\includegraphics[width=0.8\linewidth]{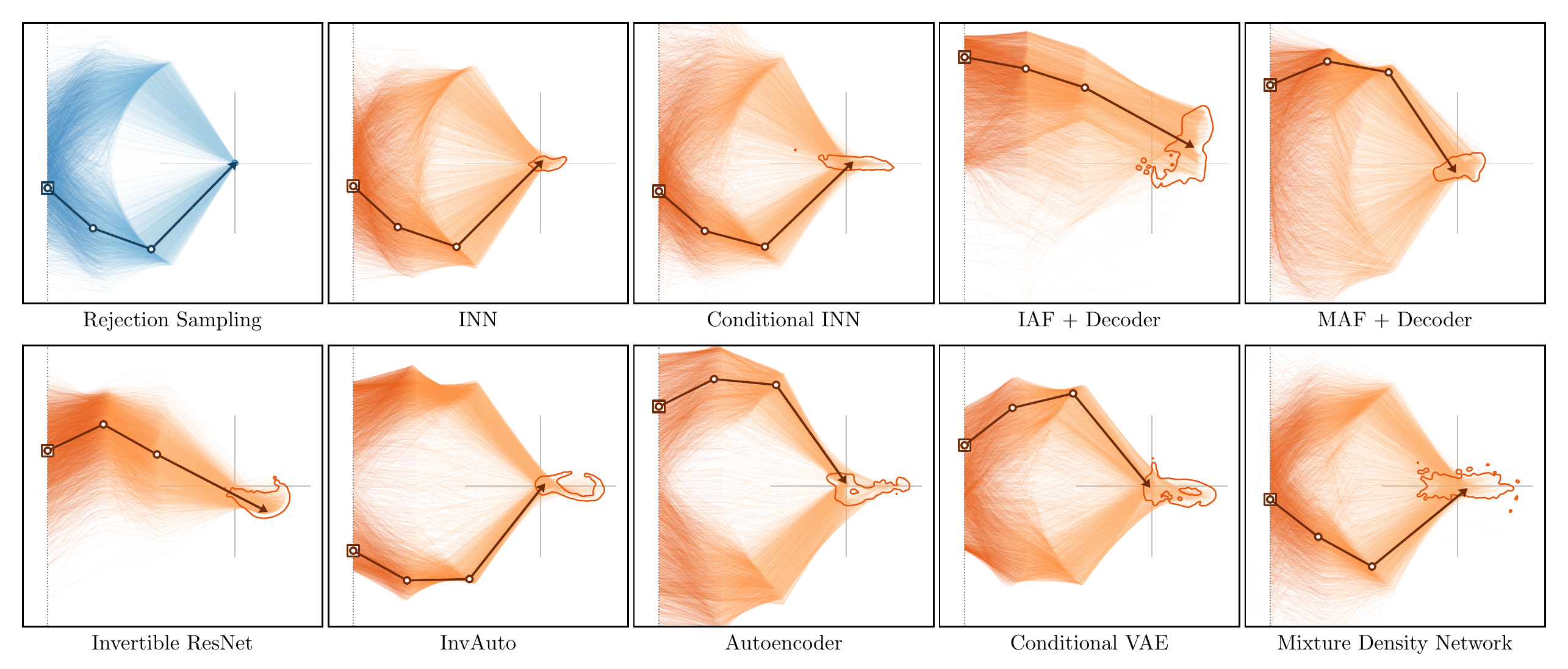}%
\vspace{-1.5em}%
\caption{
Qualitative results for the inverse kinematics benchmark.
The faint lines are arm configurations sampled from each model's predicted posterior $\hat{p}(\x \,|\, \y^*)$, the target point $\y^* = [1.5, 0]$ is indicated by a gray cross.
We emphasize the most likely arm (determined by mean shift) as a bold line.
The contour around the target marks the area containing $97\%$ of the sampled arms' end points.
}\vspace{-1em}
\label{fig:kinematics-results}
\end{figure*}

\subsection{Inverse Kinematics}
\label{inverse-kinematics-intro}
First is the geometrical example used by \cite{ardizzone2018analyzing}, which asks about configurations of a multi-jointed 2D arm that end in a given position, see \cref{fig:inverse-kinematics-prior} left.
The forward process takes a starting height $x_1$ and the three joint angles $x_2, x_3, x_4$, and returns the coordinate of the arm's end point $\y = [y_1, y_2]$ as
\begin{align*}
y_1 \!&=\! l_1 \sin(x_2) + l_2 \sin(x_2 \!+\! x_3) + l_3 \sin(x_2 \!+\! x_3 \!+\! x_4) \!+\! x_1 \\
y_2 \!&=\! l_1 \cos(x_2) + l_2 \cos(x_2 \!+\! x_3) + l_3 \cos(x_2 \!+\! x_3 \!+\! x_4)
\end{align*}
with segment lengths $l_1 = \tfrac{1}{2},\; l_2 = \tfrac{1}{2}$ and $l_3 = 1$.

Parameters $\x$ follow a Gaussian prior $\x \sim \mathcal{N}(\mathbf{0},\; \boldsymbol{\sigma}^2 \!\cdot\! \mathbf{I})$ with 
$\boldsymbol{\sigma}^2 = [\tfrac{1}{16}, \tfrac{1}{4}, \tfrac{1}{4}, \tfrac{1}{4}]$.
The inverse problem is to find the distribution $p(\x \,|\, \y^*)$ of all arm configurations $\x$ that end at some observed 2D position $\y^*$.

\subsection{Inverse Ballistics}
\label{inverse-ballistics-intro}
A similar, more physically motivated problem in the 2D plane arises when an object is thrown from a starting position $(x_1, x_2)$ with angle $x_3$ and initial velocity $x_4$.
This setup is illustrated in \cref{fig:inverse-kinematics-prior}, right.
For given gravity $g$, object mass $m$ and air resistance $k$, the object's trajectory $\mathbf{T}(t)$ can be computed as
\begin{align*}
T_1(t) &= x_1 - \frac{v_1 m}{k} \cdot \left( e^{-\tfrac{kt}{m}} - 1 \right) \\
T_2(t) &= x_2 - \frac{m}{k^2} \cdot \left( \big(\, gm + v_2 k \,\big) \cdot \left( e^{-\tfrac{kt}{m}} - 1 \right) + gtk \right)
\end{align*}
with $v_1 = x_4 \cdot \cos{x_3}$ and $v_2 = x_4 \cdot \sin{x_3}$. We define the location of impact as $y = T_1(t^*)$, where $t^*$ is the solution of $T_2(t^*) = 0$, i.e.~the trajectory's intersection with the $x_1$-axis of the coordinate system (if there are two such points we take the rightmost one, and we only consider trajectories that do cross the $x_1$-axis).
Note that here, $y$ is one-dimensional.

We choose the parameters' priors as $x_1 \sim \mathcal{N}(0,\; \tfrac{1}{4}),\; x_2 \sim \mathcal{N}(\tfrac{3}{2},\; \tfrac{1}{4}),\; x_3 \sim \mathcal{U}(9^{\circ},\; 72^{\circ})$ and $x_4 \sim \textrm{Poisson}(15)$.

The inverse problem here is to find the distribution $p(\x \,|\, y^*)$ of all throwing parameters $\x$ that share the same observed impact location $y^*$.

\section{Experiments}\vspace{-3pt}
\label{sec:experiments}
To compare all approaches in a fair setting, we use the same training data, train for the same number of batches and epochs and choose layer sizes such that all models have roughly the same number of trainable parameters (${\sim} 3\,\textrm{M}$).

We quantify the correctness of the generated posteriors in two ways, using $1000$ unseen conditions $\y^*$ obtained via prior and forward process.
Firstly, we use MMD (\cref{eq:mmd}, \citep{gretton2012kernel}) to compute the \textit{posterior mismatch} between the distribution $\hat{p}(\x \,|\, \y^*)$ generated by a model and a ground truth estimate $p_\mathrm{gt}(\x \,|\, \y^*)$ obtained via rejection sampling:
\begin{align}
 Err_\mathrm{post} &=
 \mathrm{MMD}\bigl( \hat{p}(\x \,|\, \y^*),\, p_\mathrm{gt}(\x \,|\, \y^*) \bigr)
 \label{eq:posterior-mismatch}
\end{align}
Secondly, we apply the true forward process $f$ to the generated samples $\x$ and measure the \textit{re-simulation error} as the mean squared distance to the target $\y^*$:
\begin{align}
 Err_\mathrm{resim} &= \mathbb{E}_{\,\x \sim \hat{p}(\x \,|\, \y^*)}
 \left\lVert f(\x) - \y^* \right\rVert_2^2
 \label{eq:re-simulation-error}
\end{align}
Finally, we report the inference time for each implementation using one \emph{GTX 1080 Ti}.

\subsection{Inverse Kinematics}\vspace{-3pt}
\label{inverse-kinematics-results}
Quantitative results for the kinematics benchmark are shown in \cref{tab:kinematics-results} (extra detail in \cref{fig:kinematics-boxplot}), while qualitative results for one challenging end point $\y^*$ are plotted in \cref{fig:kinematics-results}.

Architectures based on coupling layers (INN, cINN) achieve the best scores on average, followed by the simple autoencoder.
The invertible ResNet exhibits some mode collapse, as seen in \cref{fig:kinematics-results}, bottom left.
Note that we were unable to train our iResNet-implementation with the estimated Jacobian determinants, which were too inaccurate, and resorted to the loss from \cref{eq:l2-mmd}.
Similarly we would expect the autoregressive models, in particular IAF, to converge much better with more careful tuning.

MDN on the other hand performs very well for both error measures.
Note however that a full precision matrix $\boldsymbol{\Sigma}_x^{-1}$ is needed for this, as a purely diagonal $\boldsymbol{\Sigma}_x = \mathbf{I}\boldsymbol{\sigma}_x$ fails to model the potentially strong covariance among variables $x_i$.
Since $\boldsymbol{\Sigma}_x^{-1}$ grows quadratically with the size of $\x$ and a matrix inverse is needed during inference, the method is very slow and does not scale to higher dimensions.

\begin{table*}[ht!]
\caption{
Quantitative results for the inverse ballistics benchmark.
Rows and columns have the same meaning as in \cref{tab:kinematics-results}.
}
\label{tab:ballistics-results}
\begin{center}\begin{small}\begin{sc}
\begin{tabular}{lcccccc}
\toprule
Method & $Err_\mathrm{post}$ (\ref{eq:posterior-mismatch}) & $Err_\mathrm{resim}$ (\ref{eq:re-simulation-error}) & Inference in ms & dim($\z$) & ML Loss & $y$-Supervision \\
\midrule
INN & \textbf{0.047} & \textbf{0.019} & 21 & ${\bullet}{\bullet}{\bullet}$ & $\checkmark$ & $\checkmark$ \\
INN (L2 + MMD) & 0.060 & 3.668 & 21 & ${\bullet}{\bullet}{\bullet}$ & & $\checkmark$ \\
cINN & \textbf{0.047} & 0.437 & 22 & ${\bullet}{\bullet}{\bullet}{\bullet}$ & $\checkmark$ & \\
IAF + Decoder & 0.323 & 3.457 & \textbf{0} & ${\bullet}{\bullet}{\bullet}{\bullet}$ & $\checkmark$ & $\checkmark$ \\
MAF + Decoder & 0.213 & 1.010 & \textbf{0} & ${\bullet}{\bullet}{\bullet}{\bullet}$ & $\checkmark$ & $\checkmark$ \\
iResNet & 0.084 & 0.091 & 307 & ${\bullet}{\bullet}{\bullet}$ & & $\checkmark$ \\
InvAuto & 0.156 & 0.315 & 1 & ${\bullet}{\bullet}{\bullet}$ & & $\checkmark$ \\
Autoencoder & 0.049 & 0.052 & 1 & ${\bullet}{\bullet}{\bullet}$ & & $\checkmark$ \\
cVAE & 4.359 & 0.812 & \textbf{0} & ${\bullet}{\bullet}{\bullet}$ & & \\
\midrule
\textcolor{gray}{MDN} & \textcolor{gray}{0.048} & \textcolor{gray}{0.184} & \textcolor{gray}{175} & \textcolor{gray}{${\bullet}{\bullet}{\bullet}{\bullet}$} & \textcolor{gray}{$\checkmark$} & \\
\bottomrule
\end{tabular}
\end{sc}\end{small}\end{center}\vspace{-1.5em}
\end{table*}

\begin{figure*}[ht!]
\centering
\includegraphics[width=0.8\linewidth]{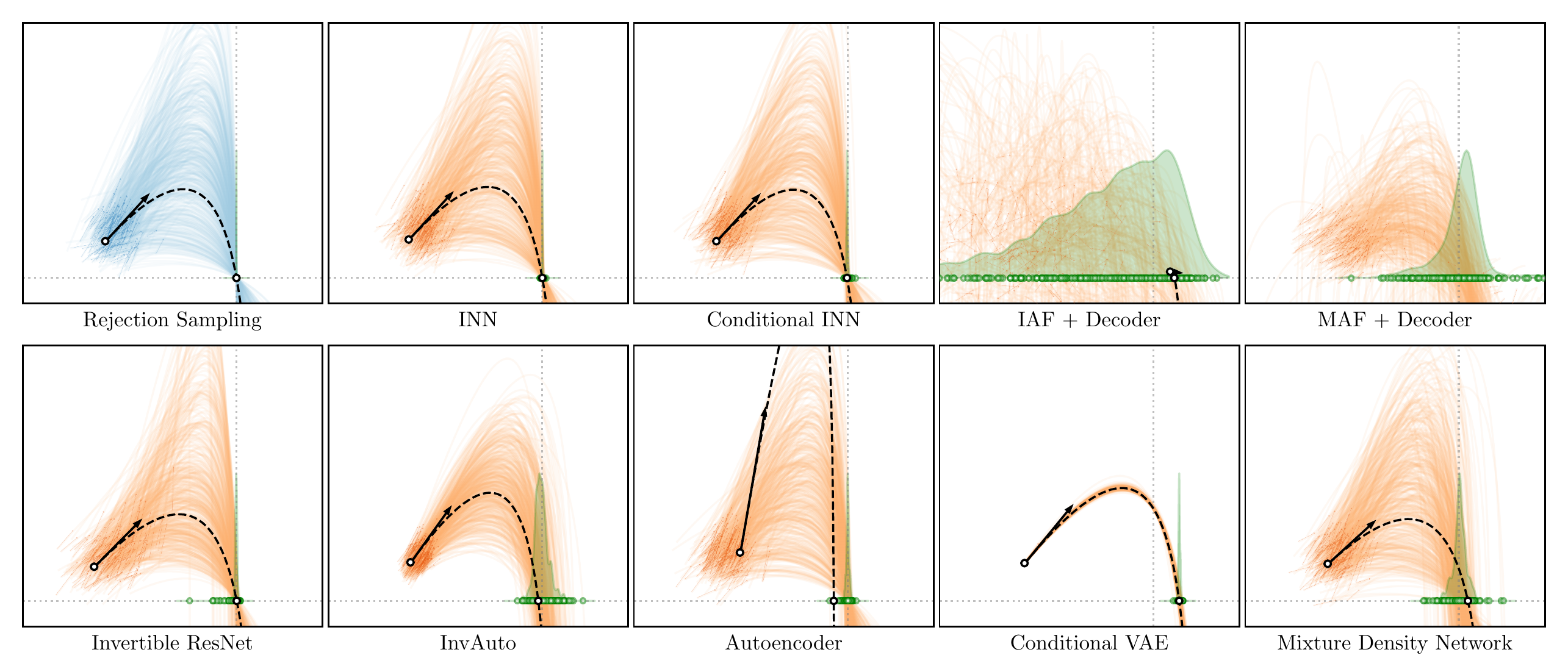}
\vspace{-1.5em}%
\caption{
Qualitative results for the inverse ballistics benchmark.
Faint lines show the trajectories of sampled throwing parameters and as above, bold is the most likely one.
A vertical line marks the target coordinate $y^* = 5$, the distribution of actual impacts is shown in green.}\vspace{-1em}
\label{fig:ballistics-results}
\end{figure*}

\subsection{Inverse Ballistics}\vspace{-3pt}
\label{inverse-ballistics-results}
Quantitative results for the ballistics benchmark are shown in \cref{tab:ballistics-results} (extra detail in \cref{fig:ballistics-boxplot}), while qualitative results for one representative impact location $y^*$ are plotted in \cref{fig:ballistics-results}.

Again we see INN, cINN and the simple autoencoder perform best.
Notably, we could not get the conditional VAE to predict proper distributions on this task; instead it collapses to some average trajectory with very high posterior mismatch.
The invertible ResNet does better here, perhaps due to the more uni-modal posteriors, but IAF and MAF again fail to capture the distributions properly at all.

Due to the presence of extreme outliers for the error measures in this task, the averages in \cref{tab:ballistics-results} are computed with clamped values and thus somewhat distorted.
\cref{fig:ballistics-boxplot} gives a better impression of the distribution of errors.
There the INN trained with \cref{eq:ml-yz} appears the most robust model (smallest maximal errors), followed by the autoencoder.
cINN and iResNet come close in performance if outliers are ignored.

\section{Discussion and Outlook} 
In both our benchmarks, models based on RealNVP \cite{dinh2016density} and the standard autoencoder take the lead, while other invertible architectures seem to struggle in various ways.
Success in our experiments was neither tied to maximum likelihood training, nor to the use of a supervised loss on the forward process.

We are aware that training of some models can probably be improved, and welcome input from experts to do so. 
In the future, the comparison should also include ODE-based methods like \citet{chen2018neural,grathwohl2018ffjord}, variants of Parallel WaveNet \cite{oord2018parallel} and classical approaches to Bayesian estimation such as MCMC. 
Ideally, this paper will encourage the community to join our evaluation efforts and possibly set up an open challenge with additional benchmarks and official leader boards.

Code for the benchmarks introduced here can be found at \url{https://github.com/VLL-HD/inn_toy_data}.

\section*{Acknowledgements}
J.~Kruse, C.~Rother and U.~K\"othe received financial support from the European Research Council (ERC) under the European Unions Horizon 2020 research and innovation program (grant agreement No 647769).
J.~Kruse was additionally supported by Informatics for Life funded by the Klaus Tschira Foundation.
L.~Ardizzone received funding by the Federal Ministry of Education and Research of Germany project `High Performance Deep Learning Framework' (No 01IH17002).



\bibliography{references}
\bibliographystyle{icml2019}

\clearpage
\appendix

\begin{minipage}[c]{2\linewidth}

\section*{Appendix}

\begin{figure}[H]
\centering
\includegraphics[width=0.95\linewidth]{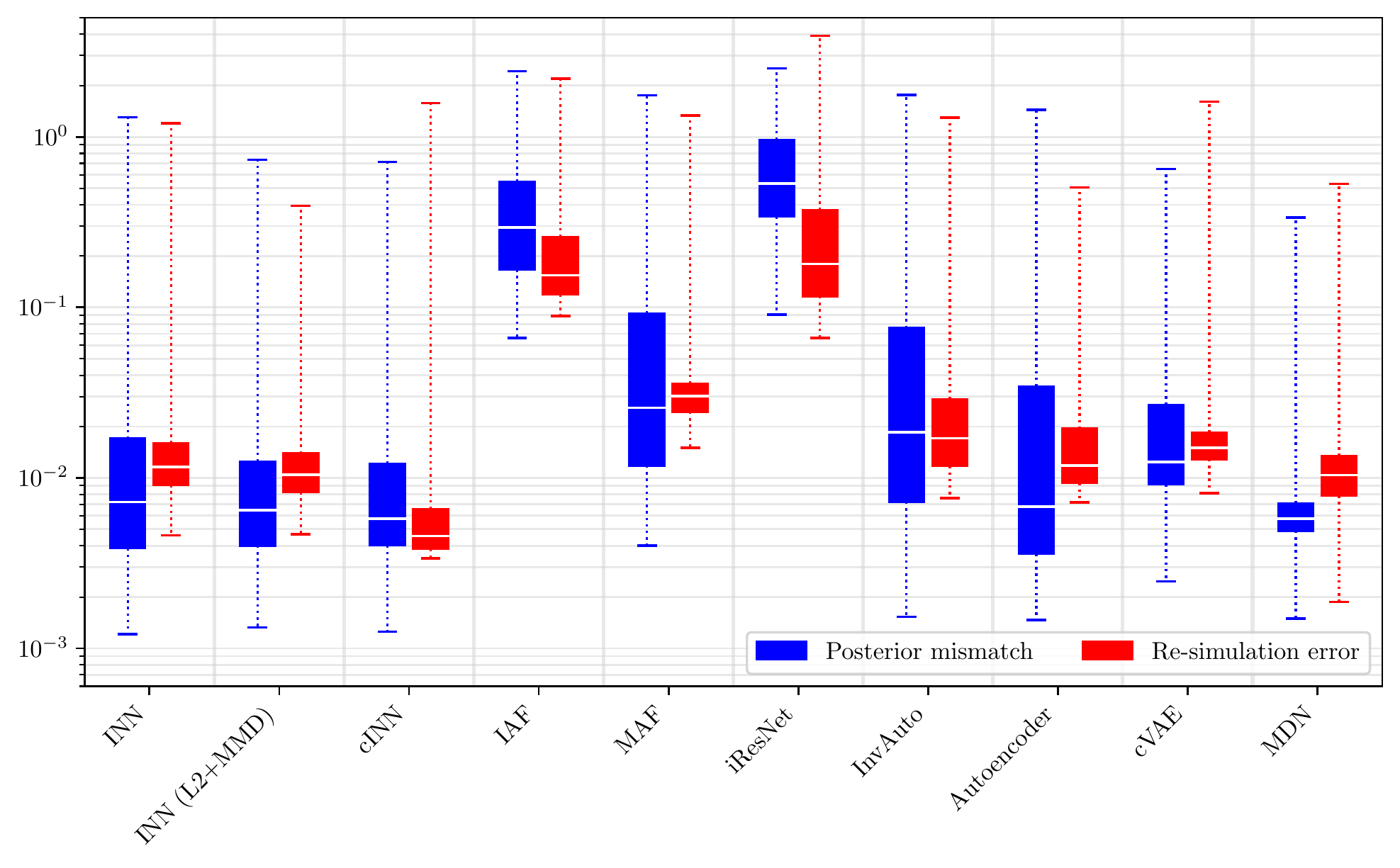}
\vspace{-1em}%
\caption{Boxplot of inverse kinematics results from \cref{tab:kinematics-results}.
The \textit{posterior mismatch} \cref{eq:posterior-mismatch} is shown in blue and the \textit{re-simulation error} \cref{eq:re-simulation-error} in red.
Boxes extend from the lower to upper quartile values of the data and a white line marks the respective median.
The dotted lines show the full range of results, including outliers.
We use log-scale to accommodate extreme values.}\vspace{-1em}
\label{fig:kinematics-boxplot}
\end{figure}

\begin{figure}[H]
\centering
\includegraphics[width=0.95\linewidth]{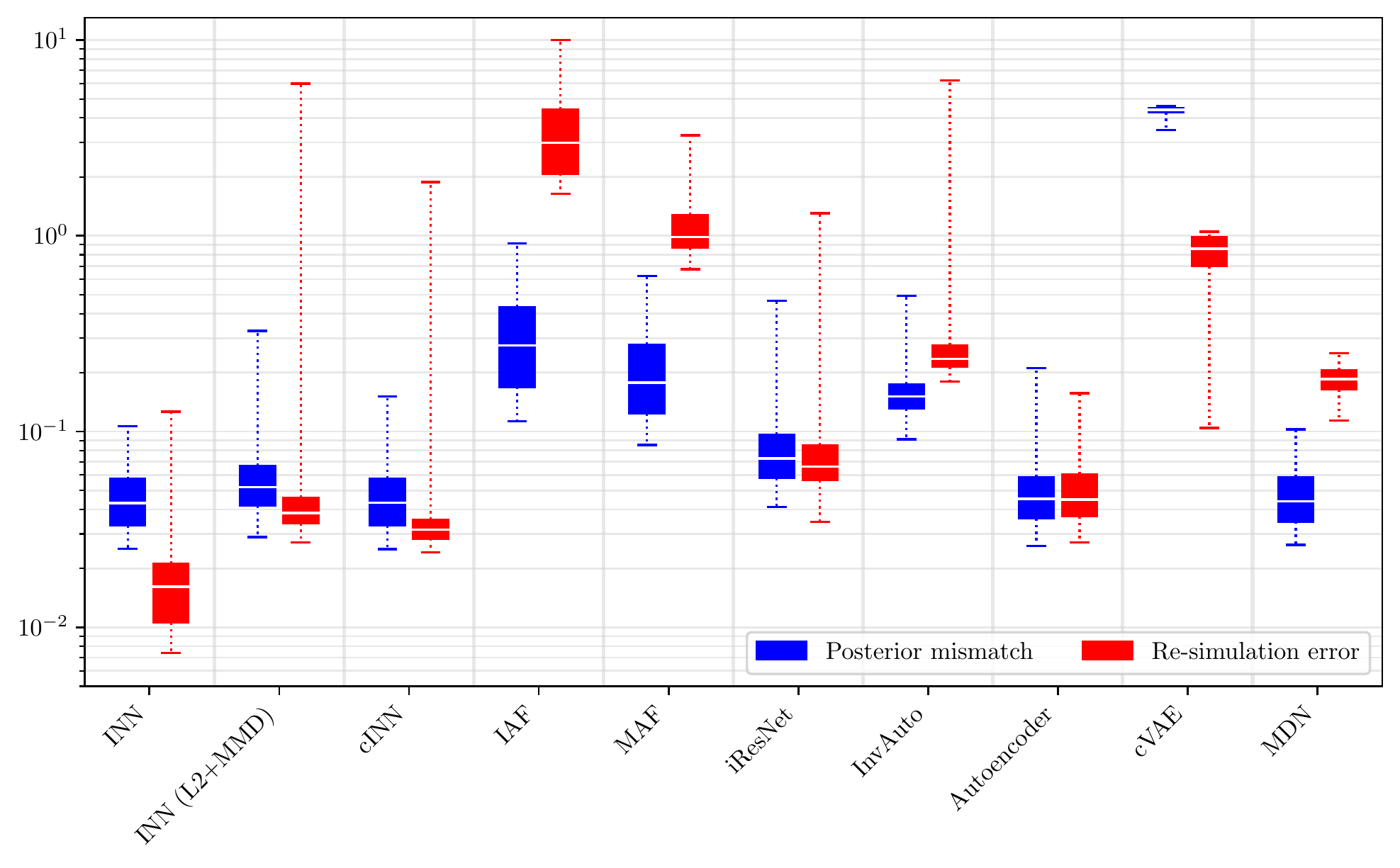}
\vspace{-1em}%
\caption{Boxplot of inverse ballistics results from \cref{tab:ballistics-results}.
The plot follows the same layout as \cref{fig:ballistics-boxplot}}\vspace{-1em}
\label{fig:ballistics-boxplot}
\end{figure}

\end{minipage}

\end{document}

%% file: figures/INN_model.tex
\begin{tikzpicture}[
    data domain/.style = {fill = black!20, draw = black, rounded corners = 3pt, minimum height = 7em, align = center, font = \Large},
    fat arrow/.style = {double arrow, fill = vll-dark, draw = black, anchor = mid, align = center, text depth = -0.5pt, text width = 11em, text = black!5},
    dashed arrow/.style = {dash pattern=on 6pt off 2pt, -{Triangle[length=9pt, width=6pt]}, shorten < = 3pt, shorten > = 2pt, line width = 2pt, draw = black!50, fill = black!50},
    arrow label/.style = {font = \scriptsize\sffamily, color = black!75, scale = 1.15},
    curly/.style = {decoration = {brace, raise = 3pt, amplitude = 5pt}, decorate},
    loss/.style = {black!75, midway, align = center, font = \footnotesize\sffamily\bfseries}]

    \node [data domain, fill = vll-green!50] (x) {$\x$};

    \node [fat arrow, right = 6pt of x.east] (INN) {\textsf{\textbf{Invertible Neural Net}}};

    \node [right = 3pt of INN.east] (yz) {
        \begin{tikzpicture}
            \path [bicolor = {vll-orange!50 and black!5}, rounded corners = 3pt, draw = black]
            (0,0) rectangle (1.5em, 7em);
        \end{tikzpicture}
    };
    \node [align = center, minimum height = 7em, font = \Large] at (yz) (yz-label) {$\y$ \\[1em] $\z$};


    \draw[curly] (x.south west) -- (x.north west) node [loss, xshift = -0.75em, rotate = 0, anchor = east, align = right] {Optional \\ MMD to \\ match prior};
    \draw[curly] (yz-label.north east) -- ([yshift=1pt] yz-label.east) node [loss, xshift = 0.75em, rotate = 0, anchor = west, align = left] {L2 to match \\ training data};
    \draw[curly] ([yshift=-1pt] yz-label.east) -- (yz-label.south east) node [loss, xshift = 0.75em, rotate = 0, anchor = west, align = left] {MMD to \\ match $\mathcal{N}(\mathbf{0}, \mathbf{1})$};

\end{tikzpicture}

%% file: figures/cINN_model.tex
\begin{tikzpicture}[
    data domain/.style = {fill = black!20, draw = black, rounded corners = 3pt, minimum height = 7em, align = center, font = \Large},
    fat arrow/.style = {double arrow, fill = vll-dark, draw = black, anchor = mid, align = center, text depth = -0.5pt, text width = 11em, text = black!5},
    dashed arrow/.style = {dash pattern=on 6pt off 2pt, -{Triangle[length=9pt, width=6pt]}, shorten < = 3pt, shorten > = 2pt, line width = 2pt, draw = black!50, fill = black!50},
    arrow label/.style = {font = \scriptsize\sffamily, color = black!75, scale = 1.15},
    curly/.style = {decoration = {brace, raise = 3pt, amplitude = 5pt}, decorate},
    loss/.style = {black!75, midway, align = center, font = \footnotesize\sffamily\bfseries}]

    \node [data domain, minimum height = 4em, fill = vll-green!50] (x) {$\x$};

    \node [fat arrow, right = 6pt of x.east] (INN) {\textsf{\textbf{Conditional INN}}};
    \fill [fill = vll-dark] ($(INN.north) + (-0.7em, 1em)$) -- ($(INN.north) + (-0.7em, -1pt)$) -- ($(INN.north) + (0.7em, -1pt)$) -- ($(INN.north) + (0.7em, 1em)$) -- cycle;
    \draw [black] ($(INN.north) + (-0.7em, 1em)$) -- ($(INN.north) + (-0.7em, -0.5pt)$);
    \draw [black] ($(INN.north) + (0.7em, 1em)$) -- ($(INN.north) + (0.7em, -0.5pt)$);
    \draw [black] ($(INN.north) + (0.7em, 1em)$) -- ($(INN.north) + (-0.7em, 1em)$);

    \node [data domain, minimum height = 4em, fill = black!5, right = 6pt of INN.east] (z) {$\z$};

    \node [data domain, fill = vll-orange!50, minimum height = 1em, minimum width = 7em, above = 1.3em of INN.north] (y) {$\y$};

    \draw[curly, opacity = 0] (x.south west) -- (x.north west) node [loss, xshift = -0.75em, rotate = 0, anchor = east, align = right, opacity = 0] {MMD to \\ match prior};
    \draw[curly] (z.north east) -- (z.south east) node [loss, xshift = 0.75em, rotate = 0, anchor = west, align = left] {maximum \\ likelihood \\ loss};

\end{tikzpicture}

%% file: figures/autoregressive_flow_model.tex
\begin{tikzpicture}[
    data domain/.style = {fill = black!20, draw = black, rounded corners = 3pt, minimum height = 7em, align = center, font = \Large},
    fat arrow/.style = {double arrow, fill = vll-dark, draw = black, anchor = mid, align = center, text depth = -0.5pt, text width = 11em, text = black!5},
    dashed arrow/.style = {dash pattern=on 6pt off 2pt, -{Triangle[length=9pt, width=6pt]}, shorten < = 3pt, shorten > = 2pt, line width = 2pt, draw = black!50, fill = black!50},
    arrow label/.style = {font = \scriptsize\sffamily, color = black!75, scale = 1.15},
    curly/.style = {decoration = {brace, raise = 3pt, amplitude = 5pt}, decorate},
    loss/.style = {black!75, midway, align = center, font = \footnotesize\sffamily\bfseries}]

    \node [data domain, minimum height = 3.3em, fill = vll-green!50] (x1) {$\x$};
    \node [data domain, minimum height = 3.3em, fill = vll-green!50, below = 0.4em of x1.south] (x2) {$\hat{\x}$};
    \coordinate (x) at ($0.5*(x1.east) + 0.5*(x2.east)$);

    \node [data domain, opacity = 0, right = 3em of x.east, minimum height = 1em, minimum width = 9em] (y) {};

    \node [right = 2.3em of y.east] (yz) {
        \begin{tikzpicture}
            \path [bicolor = {vll-orange!50 and black!5}, rounded corners = 3pt, draw = black]
            (0,0) rectangle (1.5em, 7em);
        \end{tikzpicture}
    };
    \node [align = center, minimum height = 7em, font = \Large] at (yz) (yz-label) {$\y$ \\[1em] $\z$};

    \node [fat arrow, single arrow, text width = 10em, shape border rotate = 0, above right = -3pt and -1pt of y.north, fill = vll-dark!50] (E) {\textsf{\textbf{Autoregressive flow}}};

    \node [fat arrow, single arrow, text width = 10em, shape border rotate = 180, below left = -3pt and -1pt of y.south, fill = vll-dark] (D) {\textsf{\textbf{Decoder}}};

    \draw[curly] (x2.south west) -- (x1.north west) node [loss, xshift=-1.5em, rotate = 90] {cycle loss};
    \draw[curly] (yz-label.north east) -- (yz-label.south east) node [loss, xshift = 0.75em, rotate = 0, anchor = west, align = left] {maximum \\ likelihood \\ loss};

\end{tikzpicture}

%% file: figures/iResNet_model.tex
\begin{tikzpicture}[
    data domain/.style = {fill = black!20, draw = black, rounded corners = 3pt, minimum height = 7em, align = center, font = \Large},
    fat arrow/.style = {double arrow, fill = vll-dark, draw = black, anchor = mid, align = center, text depth = -0.5pt, text width = 12em, text = black!5},
    dashed arrow/.style = {dash pattern=on 6pt off 2pt, -{Triangle[length=9pt, width=6pt]}, shorten < = 3pt, shorten > = 2pt, line width = 2pt, draw = black!50, fill = black!50},
    arrow label/.style = {font = \scriptsize\sffamily, color = black!75, scale = 1.15},
    curly/.style = {decoration = {brace, raise = 3pt, amplitude = 5pt}, decorate},
    loss/.style = {black!75, midway, align = center, font = \footnotesize\sffamily\bfseries}]

    \node [data domain, fill = vll-green!50] (x) {$\x$};

    \node [fat arrow, right = 6pt of x.east] (INN) {\textsf{\textbf{Invertible Residual Net}}};

    \node [right = 3pt of INN.east] (yz) {
        \begin{tikzpicture}
            \path [bicolor = {vll-orange!50 and black!5}, rounded corners = 3pt, draw = black]
            (0,0) rectangle (1.5em, 7em);
        \end{tikzpicture}
    };
    \node [align = center, minimum height = 7em, font = \Large] at (yz) (yz-label) {$\y$ \\[1em] $\z$};

    \path [dashed arrow] ([yshift=-7pt] x.north east) -- ([yshift=-7pt] yz-label.north west) node [arrow label, midway, above] {feed-forward with Lipschitz correction};
    \path [dashed arrow] ([yshift=7pt] yz-label.south west) -- ([yshift=7pt] x.south east) node [arrow label, midway, below] {layer by layer fixed-point iteration};

    \draw[curly] (yz-label.north east) -- ([yshift=1pt] yz-label.east) node [loss, xshift = 0.75em, rotate = 0, anchor = west, align = left] {L2 to match \\ training data};
    \draw[curly] ([yshift=-1pt] yz-label.east) -- (yz-label.south east) node [loss, xshift = 0.75em, rotate = 0, anchor = west, align = left] {MMD to \\ match $\mathcal{N}(\mathbf{0}, \mathbf{1})$};

\end{tikzpicture}

%% file: figures/invAuto_model.tex
\begin{tikzpicture}[
    data domain/.style = {fill = black!20, draw = black, rounded corners = 3pt, minimum height = 7em, align = center, font = \Large},
    fat arrow/.style = {double arrow, fill = vll-dark, draw = black, anchor = mid, align = center, text depth = -0.5pt, text width = 11em, text = black!5},
    dashed arrow/.style = {dash pattern=on 6pt off 2pt, -{Triangle[length=9pt, width=6pt]}, shorten < = 3pt, shorten > = 2pt, line width = 2pt, draw = black!50, fill = black!50},
    arrow label/.style = {font = \scriptsize\sffamily, color = black!75, scale = 1.15},
    curly/.style = {decoration = {brace, raise = 3pt, amplitude = 5pt}, decorate},
    loss/.style = {black!75, midway, align = center, font = \footnotesize\sffamily\bfseries}]

    \node [data domain, minimum height = 3.3em, fill = vll-green!50] (x1) {$\x$};
    \node [data domain, minimum height = 3.3em, fill = vll-green!50, below = 0.4em of x1.south] (x2) {$\hat{\x}$};
    \coordinate (x) at ($0.5*(x1.east) + 0.5*(x2.east)$);

    \node [fat arrow, right = 6pt of x] (INN) {\textsf{\textbf{Invertible Autoencoder}}};

    \node [right = 3pt of INN.east] (yz) {
        \begin{tikzpicture}
            \path [bicolor = {vll-orange!50 and black!5}, rounded corners = 3pt, draw = black]
            (0,0) rectangle (1.5em, 7em);
        \end{tikzpicture}
    };
    \node [align = center, minimum height = 7em, font = \Large] at (yz) (yz-label) {$\y$ \\[1em] $\z$};

    \path [dashed arrow] ([yshift=-7pt] x1.north east) -- ([yshift=-7pt] yz-label.north west) node [arrow label, midway, above] {use weights $\mathbf{W}$ and LeakyReLU};
    \path [dashed arrow] ([yshift=7pt] yz-label.south west) -- ([yshift=7pt] x2.south east) node [arrow label, midway, below] {use $\mathbf{W}^\top$ and inverse LeakyReLU};

    \draw[curly] (x2.south west) -- (x2.north west) node [loss, xshift = -0.75em, rotate = 90, anchor = south, align = center] {Optional \\ MMD};
    \draw[curly, decoration = {brace, raise = 3.25em, amplitude = 5pt}] (x2.south west) -- (x1.north west) node [loss, xshift = -3.5em, rotate = 90, anchor = south, align = center] {cycle loss};
    \draw[curly] (yz-label.north east) -- ([yshift=1pt] yz-label.east) node [loss, xshift = 0.75em, rotate = 0, anchor = west, align = left] {L2};
    \draw[curly] ([yshift=-1pt] yz-label.east) -- (yz-label.south east) node [loss, xshift = 0.75em, rotate = 0, anchor = west, align = left] {MMD};

\end{tikzpicture}

%% file: figures/AE_model.tex
\begin{tikzpicture}[
    data domain/.style = {fill = black!20, draw = black, rounded corners = 3pt, minimum height = 7em, align = center, font = \Large},
    fat arrow/.style = {double arrow, fill = vll-dark, draw = black, anchor = mid, align = center, text depth = -0.5pt, text width = 11em, text = black!5},
    dashed arrow/.style = {dash pattern=on 6pt off 2pt, -{Triangle[length=9pt, width=6pt]}, shorten < = 3pt, shorten > = 2pt, line width = 2pt, draw = black!50, fill = black!50},
    arrow label/.style = {font = \scriptsize\sffamily, color = black!75, scale = 1.15},
    curly/.style = {decoration = {brace, raise = 3pt, amplitude = 5pt}, decorate},
    loss/.style = {black!75, midway, align = center, font = \footnotesize\sffamily\bfseries}]

    \node [data domain, minimum height = 3.3em, fill = vll-green!50] (x1) {$\x$};
    \node [data domain, minimum height = 3.3em, fill = vll-green!50, below = 0.4em of x1.south] (x2) {$\hat{\x}$};
    \coordinate (x) at ($0.5*(x1.east) + 0.5*(x2.east)$);

    \node [data domain, opacity = 0, right = 2.6em of x.east, minimum height = 1em, minimum width = 7em] (y) {};

    \node [right = 2.3em of y.east] (yz) {
        \begin{tikzpicture}
            \path [bicolor = {vll-orange!50 and black!5}, rounded corners = 3pt, draw = black]
            (0,0) rectangle (1.5em, 7em);
        \end{tikzpicture}
    };
    \node [align = center, minimum height = 7em, font = \Large] at (yz) (yz-label) {$\y$ \\[1em] $\z$};

    \node [fat arrow, single arrow, text width = 8em, shape border rotate = 0, above right = -3pt and -1pt of y.north, fill = vll-dark!50] (E) {\textsf{\textbf{Encoder}}};

    \node [fat arrow, single arrow, text width = 8em, shape border rotate = 180, below left = -3pt and -1pt of y.south, fill = vll-dark] (D) {\textsf{\textbf{Decoder}}};

    \draw[curly] (x2.south west) -- (x1.north west) node [loss, xshift=-1.5em, rotate = 90] {cycle loss};
    \draw[curly] (yz-label.north east) -- ([yshift=1pt] yz-label.east) node [loss, xshift = 0.75em, rotate = 0, anchor = west, align = left] {L2};
    \draw[curly] ([yshift=-1pt] yz-label.east) -- (yz-label.south east) node [loss, xshift = 0.75em, rotate = 0, anchor = west, align = left] {MMD};

\end{tikzpicture}

%% file: figures/cVAE_model.tex
\begin{tikzpicture}[
    data domain/.style = {fill = black!20, draw = black, rounded corners = 3pt, minimum height = 7em, align = center, font = \Large},
    fat arrow/.style = {double arrow, fill = vll-dark, draw = black, anchor = mid, align = center, text depth = -0.5pt, text width = 11em, text = black!5},
    dashed arrow/.style = {dash pattern=on 6pt off 2pt, -{Triangle[length=9pt, width=6pt]}, shorten < = 3pt, shorten > = 2pt, line width = 2pt, draw = black!50, fill = black!50},
    arrow label/.style = {font = \scriptsize\sffamily, color = black!75, scale = 1.15},
    curly/.style = {decoration = {brace, raise = 3pt, amplitude = 5pt}, decorate},
    loss/.style = {black!75, midway, align = center, font = \footnotesize\sffamily\bfseries}]

    \node [data domain, minimum height = 3.3em, fill = vll-green!50] (x1) {$\x$};
    \node [data domain, minimum height = 3.3em, fill = vll-green!50, below = 0.4em of x1.south] (x2) {$\hat{\x}$};
    \coordinate (x) at ($0.5*(x1.east) + 0.5*(x2.east)$);

    \node [data domain, fill = vll-orange!50, right = 3em of x, minimum height = 1em, minimum width = 7em, text depth = 1pt] (y) {$\y$};
    \node [data domain, fill = black!5, right = 3em of y.east] (z) {$\!\!\!\begin{array}{c} \boldsymbol\mu_z \\ \boldsymbol\sigma_z \end{array}\!\!\!$};

    \node [fat arrow, single arrow, shape border rotate = 0, above left = 1.5em and -4pt of y.north, fill = vll-dark!50] (E) {\textsf{\textbf{Encoder}}};
    \fill [fill = vll-dark!50] ($(y.north) + (-2.5em, 1.6em)$) -- ($(y.north) + (-2.5em, 0.3em)$) -- ($(y.north) + (-1.1em, 0.3em)$) -- ($(y.north) + (-1.1em, 1.6em)$) -- cycle;
    \draw [black] ($(y.north) + (-2.5em, 1.54em)$) -- ($(y.north) + (-2.5em, 0.3em)$) -- ($(y.north) + (-1.1em, 0.3em)$) -- ($(y.north) + (-1.1em, 1.54em)$);
    
    \node [fat arrow, single arrow, shape border rotate = 180, below right = 1.5em and -4pt of y.south, fill = vll-dark] (D) {\textsf{\textbf{Decoder}}};
    \fill [fill = vll-dark] ($(y.south) + (1.1em, -1.6em)$) -- ($(y.south) + (1.1em, -0.3em)$) -- ($(y.south) + (2.5em, -0.3em)$) -- ($(y.south) + (2.5em, -1.6em)$) -- cycle;
    \draw [black] ($(y.south) + (1.1em, -1.54em)$) -- ($(y.south) + (1.1em, -0.3em)$) -- ($(y.south) + (2.5em, -0.3em)$) -- ($(y.south) + (2.5em, -1.54em)$);
    
    \draw[curly] (x2.south west) -- (x1.north west) node [loss, xshift=-1.5em, rotate = 90] {cycle loss};
    \draw[curly] (z.north east) -- (z.south east) node [loss, xshift=1.5em, rotate = 90] {ELBO loss};

\end{tikzpicture}

%% file: figures/MDN_model.tex
\begin{tikzpicture}[
    data domain/.style = {fill = black!20, draw = black, rounded corners = 3pt, minimum height = 7em, align = center, font = \Large},
    fat arrow/.style = {double arrow, fill = vll-dark, draw = black, anchor = mid, align = center, text depth = -0.5pt, text width = 12em, text = black!5},
    dashed arrow/.style = {dash pattern=on 6pt off 2pt, -{Triangle[length=9pt, width=6pt]}, shorten < = 3pt, shorten > = 2pt, line width = 2pt, draw = black!50, fill = black!50},
    arrow label/.style = {font = \scriptsize\sffamily, color = black!75, scale = 1.15},
    curly/.style = {decoration = {brace, raise = 3pt, amplitude = 5pt}, decorate},
    loss/.style = {black!75, midway, align = center, font = \footnotesize\sffamily\bfseries}]

    \node [data domain, minimum height = 3.3em, fill = vll-green!50] (x1) {$\x$};
    \node [data domain, minimum height = 3.3em, fill = vll-green!50, below = 0.4em of x1.south] (x2) {\scalebox{0.5}{$\!\!\!\begin{array}{c} \boldsymbol\mu_x \\[5pt] \boldsymbol\Sigma_x^{-1} \end{array}\!\!\!$}};
    \coordinate (x) at ($0.5*(x1.east) + 0.5*(x2.east)$);

    \node [fat arrow, single arrow, shape border rotate=180, right = 6pt of x.east] (INN) {\textsf{\textbf{Mixture Density Network}}};

    \node [data domain, fill = vll-orange!50, right = 3pt of INN.east] (y) {$\y$};

    \draw[curly] (x2.south west) -- (x1.north west) node [loss, xshift=-1em, align=right, anchor=east] {Maximum \\ likelihood \\ loss};

\end{tikzpicture}